\documentclass[twocolumn]{article}

\RequirePackage{amsthm,amsmath,amsfonts,amssymb}
\RequirePackage{graphicx}
\usepackage[utf8]{inputenc}
\usepackage{enumerate}
\usepackage{mathrsfs}
\usepackage{paralist}
\usepackage{color}
\usepackage{cite} 
\theoremstyle{plain}

\newtheorem{proposition}{Proposition}
\newtheorem{theorem}{Theorem}
\theoremstyle{remark}
\newtheorem{definition}{Definition}
\newtheorem{remark}{Remark}
\newtheorem{example}{Example}

\usepackage{tikz}
\usepackage{subfig}
\DeclareMathOperator{\expec}{E}
\DeclareMathOperator{\var}{Var}

\DeclareMathOperator{\Shap}{Shap}
\DeclareMathOperator{\Sob}{Sob}
\DeclareMathOperator{\Clo}{Clo}
\DeclareMathOperator{\Tot}{Tot}

\newcommand{\diff}{\mathrm{d}}

\newcommand{\sm}{\setminus}

\newcommand{\real}{\mathbf{R}}
\newcommand{\set}[1]{\{#1\}}

\newcommand{\one}[1]{\mathbf{1}\{#1\}}
\newcommand\powerset{\mathbf{2}^D}

\renewcommand\Pr{\mathrm{Pr}}

\renewcommand{\mathcal}{\mathscr}

\newcommand{\cpl}[1]{D \sm #1}

\newcommand{\XcplA}{X_{\cpl{A}}}
\newcommand{\xcplA}{x_{\cpl{A}}}

\title{A new paradigm for global sensitivity analysis}

\author{Gildas Mazo\thanks{Université Paris-Saclay, INRAE, MaIAGE, 78350,
    Jouy-en-Josas, France}}

\date{10 March 2026}

\begin{document}

\maketitle

\begin{abstract}
  It is well-known that Sobol indices, which count among the most
  popular sensitivity indices, are based on the Sobol
  decomposition. Here we challenge this construction by redefining
  Sobol indices without the Sobol decomposition. In fact, we show that
  Sobol indices are a particular instance of a more general concept
  which we call sensitivity measures. A sensitivity measure of a
  system taking inputs and returning outputs is a set function that is
  null at a subset of inputs if and only if, with probability one, the
  output actually does not depend on those inputs. A sensitivity
  measure evaluated at the whole set of inputs represents the
  uncertainty about the output.  We show that measuring sensitivity to
  a particular subset is akin to measuring the expected output's
  uncertainty conditionally on the fact that the inputs belonging to that
  subset have been fixed to random values. By considering all of the
  possible combinations of inputs, sensitivity measures induce an
  implicit symmetric factorial experiment with two levels, the
  factorial effects of which can be calculated. This new paradigm
  generalizes many known sensitivity indices, can create new ones, and
  defines interaction effects independently of the choice of the
  sensitivity measure. No assumption about the distribution of the
  inputs is required.
\end{abstract}

%%%%%%%%%%%%%%%%%%%%%%%%%%%%%%%%%%%%%%%%%%%%%%
%%%% Main text entry area:

\section{Introduction}
\label{sec:Intro}

Global sensitivity analysis is an important step in model checking,
understanding, and simplification~\cite{saltelli2008book,%
  saltelli2004sensitivity,prieurtarantola2015,da_veiga_basics_2021,%
  borgonovo_moment-independent_2016,saltelli2000sensitivityss}.  The
model, represented by a function $f$, is any system that takes inputs
and returns outputs. It can be a machine learning algorithm, an
artificial intelligence model or a more traditional mecanistic
mathematical model based on natural laws.

To perform a global sensitivity analysis one draws an input vector $X$
at random and runs the model to yield a random output $f(X)$. The
distribution of $f(X)$ represents the uncertainty about the
output. The distribution of $X$ represents the uncertainty about the
inputs. With each input (or subset of inputs) one finally associates a
number---a sensitivity index---that represents the contribution of
that input to the variability of the output.

There are many ways to associate a sensitivity index but there is one
that is most common and remains standard to this day. It is 
described next. For a comprehensive account of  possible methods we
refer to~\cite{da_veiga_basics_2021}.

\subsection{The Sobol decomposition}
\label{sec:globsensianalysisstandard}

The most common sensitivity indices are the Sobol indices. 
There are several kinds of Sobol indices. All of them
rest on the Sobol decomposition.

The Sobol decomposition~\cite{sobol1993sensitivity} is a decomposition of the random variable
$f(X)$ into as many uncorrelated random variables as there are ways to
form subsets of inputs. The \emph{Sobol index} $\Sob(A)$ associated
with subset $A$ is defined as the variance of the component associated
with $A$ in the decomposition of $f(X)$. The Sobol decomposition holds true if
the inputs are statistically independent (that is, the components of
$X$ are mutually independent). See Appendix~\ref{sec:SobolIndices} for
details.

From Sobol indices two kinds of indices are further defined.

\begin{description}
\item[Closed indices] The \emph{closed index} $\Clo(A)$ of a subset
  $A$ is defined as the sum of all Sobol indices $\Sob(B)$ with
  $B\subset A$. The index $\Clo(A)$ is maximal if and only if the
  function $f$ depends on its inputs $A$ only. It holds that
  \begin{equation}
    \label{eq:cloindexformula}
    \Clo(A) = \var\expec(f(X)|X_A),
  \end{equation}
  where $X_A$ denotes the subvector of $X$ that corresponds to $A$
  (see Section~\ref{sec:SensitivityMeasures} for clarification of the
  notation). Closed indices of singletons are used to discover the
  input that---if known---would reduce the output's uncertainty the
  most. 
\item[Total indices] The \emph{total index} $\Tot(A)$ of a subset $A$
  is defined as the sum of all Sobol indices $\Sob(B)$ such that
  $A\cap B\ne\emptyset$. The index $\Tot(A)$ is minimal (in fact,
  zero) if and only if the function $f$ does not depend on its inputs
  $A$. It holds
  \begin{equation}
    \label{eq:totindexformula}
    \Tot(A) = \expec\var(f(X)|\XcplA),
  \end{equation}
  where $D\sm A$ denotes the complement of $A$ in $D$, and $D$ denotes
  the set of all inputs. Total indices of singletons are used to fix
  unessential inputs to some arbitrary values.
\end{description}

Closed and total indices are directly related through the well-known
identity 
\begin{equation}
  \label{eq:dualityCloSob}
  \var\expec(f(X)|X_A) + \expec\var(f(X)|X_A) = \var f(X).
\end{equation}
Observe that we also have $\Clo(\set{j})=\Sob(\set{j})$ for every
singleton $\set{j}$. We refer to the textbooks cited at the beginning
of the present section for further details.

\begin{figure} \centering
  \subfloat[Standard theory]
  {\begin{tikzpicture}
      \node(Sob) at (0,0) {Sob};
      \node(Tot) at (0,-1) {Tot};
      \node(Clo) at (1.3,0) {Clo};
      \draw[->] (Sob) -- (Clo);
      \draw[->] (Sob) -- (Tot);
      \draw[<->] (Tot) -- (Clo);
    \end{tikzpicture}}
  \qquad
  \subfloat[New paradigm]
  {\begin{tikzpicture}
      \node(Sob) at (0,0) {Sob};
      \node(Tot) at (0,-1) {Tot};
      \node(Clo) at (1.3,0) {Clo};
      \draw[<-] (Sob) -- (Clo);
      \draw[<-] (Sob) -- (Tot);
      \draw[<->] (Tot) -- (Clo);
    \end{tikzpicture}}
  \caption{Standard theory defines total indices (Tot) and closed
    indices (Clo) from Sobol indices (Sob) while in the new paradigm
    the arrows are reversed: Sobol indices are defined from total or
    closed indices}
  \label{fig:oldnewparadigm}
\end{figure}

A schematic view of how the three kinds of sensitivity indices relate
to each other is depicted in the diagram on the left of
Figure~\ref{fig:oldnewparadigm}.

In some applications the inputs cannot be assumed statistically
independent~\cite{
  cournede_development_2013,
  lambert_quantization-based_2024}.
The Sobol decomposition must then be generalized~\cite{
  chastaing_generalized_2012,
  li_global_2010,
  zbMATH07512053,
  il_idrissi_hoeffding_2025,
  Li2017} or
alternative strategies be set out~\cite{
  kucherenko_estimation_2012,
  mara_variance-based_2012,
  owen_shapley_2017}.
In some other applications the analysis of the variance of the output's
distribution is too restrictive; one wishes to analyse other
characteristics~\cite{
  borgonovo2007771,
  da_veiga_global_2015,
  daveiga:hal-03108628,
  fort_new_2016,
  gamboa2018sensitivity,
  sobol_derivative_2010,
  rahman_f-sensitivity_2016,
  borgonovo_moment-independent_2016,
  ilidrissi:hal-03927476}. We refer to~\cite{da_veiga_basics_2021} for a
comprehensive account.  

\subsection{A change of paradigms?}
\label{sec:changeparadigmquestionmark}

The vector of Sobol indices can be written as a function of the vector
of closed indices. More precisely, it holds that
\begin{equation}
  \label{eq:SobfromClo}
  \Sob(A) = \sum_{B\subset A} (-1)^{|A\sm B|} \Clo(B)
  \quad (A\subset D).
\end{equation}
See, e.g.~\cite{ilidrissi:hal-03927476,liu2006estimating,mazo-tournier}. In other words, the
mapping $\Sob$ is the Möbius transform~\cite{rota_foundations_1964,
  aigner_combinatorial_1997} of the mapping
$\Clo$. It can be shown that equation~\eqref{eq:SobfromClo} is
equivalent to
\begin{equation}
  \label{eq:ClofromSob}
  \Clo(A) = \sum_{B\subset A} \Sob(B)
  \quad (A\subset D).
\end{equation}
In fact, equations~\eqref{eq:SobfromClo} and~\eqref{eq:ClofromSob} remain
equivalent even if we substitute arbitrary set functions for $\Clo$ and
$\Sob$, provided that they vanish at the empty set.
The pair of equations~\eqref{eq:SobfromClo} and~\eqref{eq:ClofromSob}
are sometimes refered to as the Möbius inversion formulas.

If Sobol indices can be rewritten as a function---the Möbius
transform---of the closed indices, then why not define closed indices
\emph{first}---for instance,  using~(\ref{eq:cloindexformula})---, and
\emph{then} define the Sobol indices as a function of the closed
indices? In other words, why not reverse the arrows as in the diagram
on the right of Figure~\ref{fig:oldnewparadigm}?

This point of view is similar to~\cite{ilidrissi:hal-03927476} (see the
discussion section there). In~\cite{ilidrissi:hal-03927476}, the objective
was to find decompositions of arbitrary quantities of interest of the
output's distribution in a manner similar to the Sobol decomposition;
that is, to decompose the quantity into as many components as there
are subsets of inputs in such a way that each component is related to
a particular input vector. For instance, if the quantity of interest
is the variance, then one would set $\Clo(A)$ to the variance of
$\expec(f(X)|X_A)$ and define the Sobol indices
with~(\ref{eq:SobfromClo}); the Möbius inversion formulas would then
yield the decomposition
\begin{equation*}
  \var f(X) = \sum_{A\subset D} \Sob(A).
\end{equation*}
(Take $A=D$ in~(\ref{eq:ClofromSob}) and observe that $\Clo(D)=\var
f(X)$.) Conditions on the construction of ``good'' $\Clo(A)$ to yield
decompositions with certain properties are discussed
in~\cite{ilidrissi:hal-03927476}.

One major advantage of the new paradigm described above is that
 we no longer need the Sobol decomposition to define 
sensitivity indices. This implies in particular that
\begin{itemize}
\item we would no longer need to assume statistically independent
  inputs, and,
\item we would be able to consider uncertainty measures other than the variance.
\end{itemize}

Thus, changing paradigms would allow us to address important questions
of the field.

Even if we are fine with taking the
variance as the uncertainty measure and assuming statistical
independence of the inputs, we shall see that the new point of view
is a powerfull tool to think about the problem of sensitivity
analysis in general.

\subsection{Open questions}
\label{sec:remainingquestions}

The change of paradigms described above is based on the following
reasoning: \emph{if} we want a decomposition of the variance of the
model's output (or another quantity of interest) \emph{then} we can
use the Möbius transform and the Möbius inversion formulas to get it.

But \emph{why} do we want such a decomposition in the first place? In other words, how do
we interpret~(\ref{eq:SobfromClo})? What is exactly what we are doing?
Indeed, it is not immediately obvious when we look at the formula.
After all, as noticed in~\cite{ilidrissi:hal-03927476}, one's ability
to get such a decomposition does not mean that it is meaningful. (Take
$\Clo(D)=\var f(X)$ and $\Clo(A)$ anything.)

A second weakness in the above reasoning is the central role played by the indices $\Clo$. We could as
well take $\Tot$ instead. We would then define it
with~(\ref{eq:totindexformula}), and define Sobol indices as the
Möbius transform of $\Tot$ (and not $\Clo$). This is noticed
in~\cite{mazo-tournier}.

In fact, there is one
argument to use $\Tot$ instead of $\Clo$:  total
indices---not closed indices---agree with the concept of
sensitivity in the sense that $\Tot(A)$ is null if and only if $f$ is
unsensitive to $A$. What's more, $\Tot$ and $\Clo$ are related
through~\eqref{eq:dualityCloSob} anyway.

\subsection{Main contributions}
\label{sec:maincontributions}

We take up, formalize and generalize this idea of ``reversing the
arrows'' pictured in the diagram on the right of
Figure~\ref{fig:oldnewparadigm}.

As suggested in Section~\ref{sec:remainingquestions}, we take
total indices instead of closed indices as the starting point. 

To generalize total indices, a characteristic property that we can
posit as a definition must be sought. Thus we define a
\emph{sensitivity measure} $\tau$ as a mapping that takes a subset $A$
and returns a nonnegative number $\tau(A)$ such that $\tau(A)=0$ if
and only if the function $f$ does not depend on its inputs $A$.

A general class of sensitivity measures is constructed from the
conditional probabilities of the model's output given each
subset of the input vector. A sufficient condition for $\tau(A)$ to be
a sensitivity measure is that it is the
expectation of a functional of the conditional distribution of $f(X)$
given $\XcplA$ that vanishes if and only if that conditional distribution
is a Dirac measure.
Particular choices of that functional yield particular sensitivity
measures. We exhibit several examples that lead to various sensitivity
indices of the literature.

By setting the inputs' distribution to one of the possible conditional
distributions of $X$ given $X_A$ ($A\subset D$), we show that every
sensitivity measure induces a ``thought'' symmetric factorial
experiment at two levels in which the output's uncertainty is measured
conditionally on each possible subset of inputs.  The sensitivity
measure of a given subset is then the expected outcome of
the induced factorial experiment.

To measure the effects of the inputs' uncertainties the factorial
effects of the induced factorial experiment are calculated.  We also
define weighted factorial effects by multiplying each marginal
contribution of a subset with a given weight. The common Sobol indices
are shown to coincide with these weighted factorial effects under a
certain choice of the weights and the sensitivity measure.

The rest of the paper is organized as follows. Sensitivity measures
are introduced in Section~\ref{sec:SensitivityMeasures}. The
interpretation of them as the expected outcome of a symmetric
factorial experiment with two levels in which one measures the
conditional output's uncertainty is given in
Section~\ref{sec:causes}. The calculation
of factorial effects and weighted factorial effects is given in Section~\ref{sec:uncertaintyeffects}. The link with the
traditional Sobol indices is made in
Section~\ref{sec:Sobolnewpdgm}. Section~\ref{sec:conclusion}
discusses the implications of the present work. 

%%%%%%%%%%%%%%%%%%%%%%%%%%%%%%%%%%%%%%%%%%%%%%%%%%%%%%%%%%%%%%%%%%%%%
\section{Sensitivity measures}
\label{sec:SensitivityMeasures}

Sensitivity measures as a generalization of the total index are
defined here. They are written as a functional of the conditional
distributions of the model's output given each subvector of
inputs. Two principles to construct that functional are given. We also
give examples relating sensitivity measures to well-known indices of
the literature.

\subsection{Definitions}
\label{sec:sensimeasdefinition}

Let $f$ be a (Borel-measurable) function from the $d$-dimensional
Euclidean space $\real^d$ into the real line $\real$. Let $X$ be a
random vector in $\real^d$. Let $D$ denote the index set
$\{1,\dots,d\}$.  If $A\subset D$ then $X_A\in\real^{|A|}$ denotes the
subvector of $X$ with components those of $X$ indexed by $A$. For
instance if $d=4$ and $A=\{3,1,4\}$ then $X_A = (X_1,X_3,X_4)$. If
$A=D$ then $X_D=X$. If $A=\emptyset$ then $X_{\emptyset}$ denotes an
arbitrary constant. Let $\powerset$ denote the power set of $D$, that
is, the set of all of its subsets.

A set $A\subset D$ is said to be \emph{superfluous} if there is a
(Borel-measurable) function $g$ such that $f(X) = g(X_{\cpl A})$ with
probability one.  In this case we also say that $f$ is
\emph{unsensitive} to $A$. Otherwise, $A$ is said to be
\emph{non-superfluous}, and $f$ \emph{sensitive} to $A$.  Note that
$\emptyset$ is superfluous (take $g=f$) and $D$ is non-superfluous
unless $f$ is constant. Note also that $g(\XcplA)$ must be (a version
of) the conditional expectation of $f(X)$ given $\XcplA$. Finally note
that if $A\subset B$ and $B$ is superfluous then so must be $A$.

\begin{example}
  \label{ex:octha}
  Let $f$ be the function on $\real^4$ defined by $f(x_1,x_2,x_3,x_4)
  = \sin(x_1) + x_2x_3$. Note that $x_4$ is missing, and hence
  $\set{4}$ will be superfluous for every distribution of $X$. Now
  suppose that $X_1$ is uniformly distributed on the set
  $\{-\pi,0,\pi\}$, and that $(X_2,X_3)$ is uniformly distributed on the
  set $\{(-1,1), (1,-1)\}$. Then $\{1,2,3,4\}$ is superfluous. 
\end{example}

Example~\ref{ex:octha} illustrates that for a fixed function $f$
whether or not a subset is superfluous depends on the distribution of
$X$. In the present work we consider $f$ and $X$'s distribution fixed,
and hence we keep using the word ``superfluous'' without mentioning
the function or the distribution.

The same comment applies to sensitivity measures, defined next.

\begin{definition}[sensitivity measure]\label{def:SensiMeas}
  A map $\tau: \powerset \to [0,\infty)$ is a sensitivity measure if,
  for every $A\subset D$, it holds that $\tau(A) = 0$ if and only if
  $A$ is superfluous.
\end{definition}

The number $\tau(A)$ measures the sensitivity of $f$ to its inputs
indexed by $A$. Note
that $\tau(\emptyset) = 0$. The number $\tau(D)$ is called \emph{the total
  sensitivity of $f$}, or simply \emph{the sensitivity of $f$}. It
represents the uncertainty about the output of the function $f$. 

\subsection{A general class of sensitivity measures}
\label{sec:construction}

Sensitivity measures can be constructed by comparing each conditional
distribution of $f(X)$ given $\XcplA$ with a Dirac measure.

A functional $\phi(Q)$ of a probability measure $Q$ on $\real$ will be
called a \emph{Dirac test} if $\phi(Q)\ge 0$, and, $\phi(Q)=0$ if and
only if $Q$ is a Dirac measure. Let $\phi_A(\XcplA)$ denote the
functional $\phi$ evaluated at the conditional distribution of $f(X)$
given $\XcplA$, which may be written
$\phi_A(\XcplA)=\phi(\Pr\{f(X)\in\cdot|\XcplA\})$. By abuse of
language we also call $\phi_A(\XcplA)$ a Dirac test.

\begin{theorem}
  \label{thm:fundamentaltheorem}
  If $\phi_A(\XcplA)$ is a Dirac test then the mapping defined by
  \begin{equation*}
    \tau(A) = \expec \phi_A(\XcplA)
  \end{equation*}
  is a sensitivity measure.
\end{theorem}

A sketch of proof is as follows. If $\tau(A)$ is null then
$\phi_A(\XcplA)$ must be null as well and hence, since
$\phi_A(\XcplA)$ is a Dirac test applied to the conditional
distribution of $f(X)$ given $\XcplA$, the latter distribution must a
Dirac measure, implying that $f(X)$ must be a function of $\XcplA$
only and hence that $A$ is superfluous. The detailed proof is in the
Supplementary Material (Mazo~(2026)).

Thus, to know whether $A$ is superfluous is tantamount to knowing
whether the conditional distribution of $f(X)$ given the complementary
subvector $\XcplA$ concentrates on a single point, meaning that the
model's output is certain if we were to know the value of
$\XcplA$.  

As obvious as it may seem, Theorem~\ref{thm:fundamentaltheorem} is
foundational.  It forms the basis which the concepts and results to be
developed next stem from (conditional sensitivity measures, induced
factorial experiments, and factorial effects).

\subsection{Examples}
\label{sec:sensimeasuresexamples}

Let
$\rho(y,y')$ be a nonnegative function of two reals $y$ and $y'$ such
that $\rho(y,y')=0$ if and only if $y=y'$.

A first example is given by
\begin{equation*}
  \phi_A(\XcplA) =
  \expec\left(\rho(f(V,\XcplA),f(W,\XcplA))|\XcplA\right),
\end{equation*}
where $V$ and $W$ are two random vectors in $\real^{|A|}$ such that
$V$ and $W$ are conditionally independent given $\XcplA$, and,
$(V,\XcplA)$ and $(W,\XcplA)$ have the same joint distribution as
$X$.

For the function $\rho$ we can take for instance
$\rho(y,y')=(y-y')^2/2$, leading to
$\phi_A(\XcplA)= \var(f(X)|\XcplA)$ and hence
$\tau(A)=\Tot(A)$ as seen in~(\ref{eq:totindexformula}).

Assuming for simplicity that the distribution of $f(X)$ is absolutely
continuous with respect to the Lebesgue measure, a second possibility
for $\rho$ is to take
\begin{equation*}
  \rho(y,y') = \frac{1}{2} \Pr\set{
  \min\set{y,y'} < f(\widetilde{X}) \le \max\set{y,y'}},
\end{equation*}
where here $\widetilde{X}$ is an independent copy of $X$.
Put $Y=f(V,\XcplA)$ and $Y'=f(W,\XcplA)$.
Since $\rho(Y,Y')$ is a version of
\begin{equation*}
  \frac{1}{2}\Pr\left\{\min\{Y,Y'\}<
    f(\widetilde{X})\le \max\{Y,Y'\}|\XcplA,V,W\right\},
\end{equation*}
we have that 
\begin{multline*}
  \phi_A(\XcplA) = \\
  \frac{1}{2}\Pr\left\{\min\{Y,Y'\}<
  f(\widetilde{X})\le \max\{Y,Y'\}|\XcplA\right\},
\end{multline*}
and hence
\begin{align*}
  \tau(A)
  &=\frac{1}{2} \Pr\set{\min\set{Y,Y'} < f(\widetilde{X}) \le
    \max\set{Y,Y'}}\\
  &=\frac{1}{2}( \Pr\set{Y < f(\widetilde{X}) \le Y'}
    + \Pr\set{Y' < f(\widetilde{X}) \le Y} )\\
  &=\Pr\set{Y < f(\widetilde{X}) \le Y'}\\
  &=\expec\Pr\set{Y < f(\widetilde{X}) \le Y'|\widetilde{X},\XcplA}\\
  &=\expec\left(\Pr\set{Y < f(\widetilde{X})|\widetilde{X},\XcplA}\right.\\
  &\qquad\times\left.
    \Pr\set{f(\widetilde{X})\le Y'|\widetilde{X},\XcplA}\right)\\
  &=\expec\var(\one{Y<f(\widetilde{X})}|\widetilde{X},\XcplA)\\
  &=\expec\var(\one{f(X)<f(\widetilde{X})}|\widetilde{X},\XcplA).
\end{align*}
Above $\one{y\le y'}$ denotes
the indicator function of the set $\set{y\le y'}$.
The sensitivity index
$\expec\var(\one{f(X)<f(\widetilde{X})}|\widetilde{X},\XcplA)$ was
proposed in~\cite{gamboa2018sensitivity}; it is 
 related to Chatterjee's correlation
coefficient~\cite{Chatterjee02102021}.   

A second example with a different flavor draws inspiration
from~\cite{fort_new_2016}. Let
\begin{equation*}
  \phi_A(\XcplA) =
  \min_{y\in\real} \expec\left[\rho(y,f(X))|\XcplA\right].
\end{equation*}
This and
\begin{equation*}
  \rho(y,y') = (y-y') (\alpha - \one{y\le y'})
  \quad (0<\alpha<1)
\end{equation*}
lead to a version of the quantile-based indices
of~\cite{fort_new_2016}.

\subsection{Construction of Dirac tests}
\label{sec:construction}

The examples of $\phi_A(\XcplA)$ given in
Section~\ref{sec:sensimeasuresexamples} come from the application of
two principles to construct the functional $\phi$. These are given below.

To test whether a distribution $Q$ is a dirac measure we compare $Q$
to the Dirac measure at $y$, denoted by $\delta_y$, for every possible
constant $y$. Let $\zeta(y,Q)$ be a nonnegative function of $y$ and
$Q$ such that $\zeta(y,Q)=0$ if and only if $Q=\delta_y$.

\begin{proposition}
  \label{prop:Diractest}
  Each of the functionals
  \begin{enumerate}[(i)]
  \item\label{it:firstprinciple}
    $\phi(Q) = \int_{\real} \zeta(y,Q) Q(\diff y)$, and,
  \item\label{it:secondprinciple}
    $\phi(Q) = \min_{y\in\real} \zeta(y,Q)$,
  \end{enumerate}
  is a Dirac test.

  \begin{proof}
    To prove~\eqref{it:firstprinciple},
    put $H=\{y:Q=\delta_y\}$. If $\phi(Q)=0$ then
    $Q\{y:Q\ne\delta_y\}=0$ and hence $Q(H)=1$. This implies that $H$
    is nonempty. If $y,y'\in H$ then $Q\{y\}=Q\{y'\}=1$ and hence
    $Q=\delta_y=\delta_{y'}$ and therefore $y=y'$. Thus $H$ is a
    singleton $\{y\}$ and $Q=\delta_y$. Conversely if the latter
    equality holds then by definition of $\zeta$ we have
    $0=\zeta(y,Q)=\int\zeta(y',Q)\delta_y(\diff y')=\int\zeta(y',Q)
    Q(\diff y')=\phi(Q)$.
    To prove~\eqref{it:secondprinciple}, note that
    if $\phi(Q)=0$ then $\zeta(y,Q)=0$ for some $y$ and hence
    $Q=\delta_y$. Conversely if $Q=\delta_y$ then
    $\phi(Q)=\min_{y'}\zeta(y',Q)\le\zeta(y,Q)=0$.
  \end{proof}
\end{proposition}

For the function $\zeta$ we give two examples.

\begin{example}
  \label{ex:examplezetaone}
  Set $\zeta(y,Q)=\psi(\delta_y,Q)$ where $\psi$ is some suitable
  divergence between probability measures.
\end{example}

\begin{example}
  \label{ex:examplezetatwo}
  Set 
  \begin{equation*}
    \zeta(y,Q) = \int_{\real} \rho(y,y') Q(\diff y')
  \end{equation*}
  with $\rho(y,y')$ a nonnegative function of $y$ and $y'$ such that
  $\rho(y,y')=0$ if and only if $y=y'$. If $\zeta(y,Q)=0$ then
  $\rho(y,y')=0$ for $Q$-almost all $y'$ meaning that
  $1=Q\{y':y'=y\}=Q\{y\}$, and hence $Q=\delta_y$. Conversely if
  $Q=\delta_y$ then $\zeta(y,Q)=\int \rho(y,y') \delta_y(\diff y') =
  \rho(y,y)=0$. 
\end{example}

Choosing $\zeta$ as in Example~\ref{ex:examplezetatwo} and applying
Proposition~\ref{prop:Diractest} leads to the formulas of
$\phi_A(\XcplA)$ given in Section~\ref{sec:sensimeasuresexamples}.

\section{From sensitivity measures to their causes}
\label{sec:causes}

In the previous section we said that $\tau(A)$ measures the
sensitivity of $f$ to $A$ and that $\tau(D)$ measures the output's
uncertainty.  We now would like to claim that $\tau(A)$ measures the
output's uncertainty \emph{caused} by $A$.

To this aim we introduce the method of conditioning and show that it
induces an implicit factorial experiment to analyse the output's
uncertainty. The connection with sensitivity measures is then made.

\subsection{The method of conditioning}
\label{sec:conditioning}

In practice it is of interest to study changes in the output's
uncertainty caused by increased or decreased knowledge about the
inputs. Insight can be gained by fixing a subset of the inputs and
observing the corresponding effect on the output's uncertainty.

Suppose that all components of $X$ are fixed to some
$x\in\real^d$. We can think of $x$ as the true value of the unknown
parameters of the model $f$. Since all inputs are fixed, the output's
uncertainty is null.

Now suppose that the components indexed by $A\subset D$ are in fact
uncertain, and hence are drawn at random. The distribution according
to which they are drawn is, of course, the conditional distribution of
$(X_A,\xcplA)$ given $\XcplA=\xcplA$. 

In this new setting, we still can measure
the sensitivity of the function $f$ in the same manner as before.
We still see $f$ as a function of $\real^d$. The sole difference
between Section~\ref{sec:SensitivityMeasures} and the present setting
is that the distribution according to which the inputs are drawn has
changed. (This new setting is valid since nothing was imposed on the
input distribution in Section~\ref{sec:SensitivityMeasures}.)

A mathematical description of the above idea is given next.

\begin{definition}[conditional sensitivity measure]
  A nonnegative random variable $\tau(B|\XcplA)$ is a conditional
  sensitivity measure of $B$ given $\XcplA$ if
  \begin{enumerate}[(i)]
  \item $\tau(B|\XcplA)$ is a function of (technically, measurable with
    respect to) $\XcplA$;
  \item $\tau(B|\XcplA)=0$ if and only if $A\cap B$ is superfluous.
  \end{enumerate}
  The mapping $B\mapsto\tau(B|\XcplA)$ is called a conditional
  sensitivity measure given $\XcplA$.
  If $\tau(B|\XcplA)=0$ then $B$
  is said to be conditionally superfluous given $\XcplA$.
\end{definition}

A conditional sensitivity measure of $B$ given $\XcplA$ represents the
sensitivity of $f$ to $B$ when the inputs are drawn according to the
conditional distribution of $X$ given $\XcplA$.  Observe that $D\sm A$
is conditionally superfluous given $\XcplA$. (We might say that it is
conditionally superfluous given itself.)  We call $\tau(D|\XcplA)$
\emph{the sensitivity of $f$ conditionally on $\XcplA$}. It represents
the output's uncertainty conditionally on the fact that we know $\XcplA$.

\subsection{An implicit factorial experiment}
\label{sec:implicitfactorialexper}

To find the causes of the output's uncertainty, it suffices to
``free'' every possible subvector of inputs and observe the
corresponding effect on the output's uncertainty. Since the default
state of uncertainty is 0 and is given by $\tau(D|X)$ (all inputs are
fixed), each number $\tau(D|\XcplA)$ is equal to
$\tau(D|\XcplA)-\tau(D|X)$ and thus represents the activation of $A$'s
uncertainty. We are thus in a sitation in which we do a measurement
(the output's uncertainty) under various scenarios involving all
possible combinations of some binary factors (the
activation/inhibitation of the inputs' uncertainties): this is
precisely the definition of a factorial experiment.  This factorial
experiment has two levels and is symmetric. See
Table~\ref{tab:factorialexperiment} for an illustrative example with
$d=3$. (We might say a ``thought'' factorial experiment since of
course there is no actual physical experiment.)

\begin{table}[ht]
  \centering
  \begin{tabular}{rcl}
    output's uncertainty & factors & input's distribution\\
    0 & 000 & $\Pr\{X\in\cdot|X\}=\delta_X$\\
    $\tau(D|X_{\{1,2\}})$ & 001 & $\Pr\set{X\in\cdot
                                 |X_1,X_2}$\\
    $\tau(D|X_{\{1,3\}})$ & 010 & $\Pr\set{X\in\cdot
                                 |X_1,X_3}$\\
    $\tau(D|X_{\{1\}})$ & 011 & $\Pr\set{X\in\cdot
                                 |X_1}$\\
    $\tau(D|X_{\{2,3\}})$ & 100 & $\Pr\set{X\in\cdot
                                 |X_2,X_3}$\\
    $\tau(D|X_{\{2\}})$ & 101 & $\Pr\set{X\in\cdot
                                 |X_2}$\\
    $\tau(D|X_{\{3\}})$ & 110 & $\Pr\set{X\in\cdot
                                 |X_1}$\\
    $\tau(D|X_{\emptyset})$ & 111 & $\Pr\set{X\in\cdot}$
  \end{tabular}
  \caption{Factorial experiment to study the uncertainties'
    activations when $d=3$. The $j$th factor is 1 if $\set{j}$'s
    uncertainty is active, and 0 otherwise. The symbol $\delta_X$
    denotes the Dirac measure at $X$.}
  \label{tab:factorialexperiment}
\end{table}

From the induced factorial experiment we can calculate various effects.
For instance, the number $\tau(D|X_A) - \tau(D|X_{\emptyset})$
measures the effect of the inhibition of $A$'s uncertainty, while
$\tau(D|\XcplA) - \tau(D|X)$ measures the effect of the activation of
$A$'s uncertainty. In general for $A$ and $B$ disjoint, the number
$\tau(D|X_{D\sm(A\cup B)}) - \tau(D|X_{D\sm B})$ measures the effect
of activating $A$'s uncertainty while $B$'s was active, and
$\tau(D|X_{D\sm B}) - \tau(D|X_{D\sm(A\cup B)})$ is the effect of
inhibiting $A$'s uncertainty while $B$'s was active.

\subsection{Back to sensitivity measures}
\label{sec:backtosensimeasures}

The induced factorial experiment of
Section~\ref{sec:implicitfactorialexper} is now linked to the
sensitivity measures of
Section~\ref{sec:SensitivityMeasures}. Conditional sensitivity
measures were only an intermediate.

\begin{theorem}
  \label{thm:conditional}
  If $\phi_A(\XcplA)$ is a Dirac test and
  $\tau(A)=\expec\phi_A(\XcplA)$ is its associated sensitivity measure
  then 
  \begin{equation}
    \label{eq:conditionalsensimeas}
    \tau(B|\XcplA) =
    \expec(\phi_{A\cap B}(X_{(\cpl{B})\cap A},\XcplA)|\XcplA)
  \end{equation}
  is a conditional sensitivity measure of $B$ given $\XcplA$ and
  \begin{equation}
    \label{eq:sensimeas:fundamentalform}
    \tau(A)=\expec\tau(D|\XcplA).
  \end{equation}
\end{theorem}

The proof is similar to that of Theorem~\ref{thm:fundamentaltheorem}.

A sensitivity measure, therefore, induces a conditional sensitivity
measure and hence a symmetric factorial experiment at two levels of
which it is the expected outcome.

Thus, for instance, we have that the number
  $\tau(A\cup B) - \tau(B)$
measures the average effect of activating $A$'s uncertainty while
$B$'s was active, and the number
  $\tau(B) - \tau(A\cup B)$
measures the average effect of inhibiting $A$'s uncertainty while
$B$'s was active.

\section{(Weighted) factorial effects}
\label{sec:uncertaintyeffects}

We now reap the benefits of the connection between sensitivity
measures and factorial experiments. First, factorial effects and their
generalization the weighted factorial effects are defined and their
formulas given. Then comes a symbolic illustration.

\subsection{Definitions and formulas}
\label{sec:formulasofeffects}

Among the tools to study factorial experiments, the calculation of
factorial effects is an important one~\cite{wu_experiments_2021,%
  yates_design_1937,mukerjee_modern_2006}. We generalize
slightly by considering weighted factorial effects. We shall see in
Section~\ref{sec:Sobolnewpdgm} that this will allow us to recover the
usual Sobol indices.

Let $\Delta^i$, $i=1,\dots,d$, denote the differential operator acting
on the map $\tau:\powerset\to\real$ and defined by
\begin{equation*}
  \Delta^i\tau(A) = \tau(A\cup\set{i}) - \tau(A)
  \qquad (A \subset D).
\end{equation*}
The \emph{factorial effect} of $\{i_1,\dots,i_n\} \subset D$
is defined as
\begin{multline}\label{eq:def-interactions-weights}
  I(\set{i_1,\dots,i_n})
  =\\ \sum_{A \subset D\sm\set{i_1,\dots,i_n}}
  p_{\set{i_1,\dots,i_n}}(A)
  (\Delta^{i_n} \cdots \Delta^{i_1} \tau)(A),
\end{multline}
where $\Delta^{i_n} \cdots \Delta^{i_1}$ is the composition of the
differential operators $\Delta^{i_1}, \dots, \Delta^{i_n}$, and
the numbers $p_{\set{i_1,\dots,i_n}}(A)$ ($A\subset\cpl{\set{i_1,\dots,i_n}}$) are weights
such that 
\begin{equation}\label{eq:weights}
  \left\{\begin{array}{rl}
           &p_{\set{i_1,\dots,i_n}}(A)\ge 0\\
           &\sum_{A \subset D \setminus \set{i_1,\dots,i_n}}
             p_{\set{i_1,\dots,i_n}}(A) = 1.
         \end{array}\right.
     \end{equation}
If $n=0$ then, by convention, it holds
\begin{equation*}
  I(\emptyset) = \sum_{A\subset D} p_{\emptyset}(A)\tau(A).
\end{equation*}

\begin{remark}
  Note that composition is commutative, that is,
  \begin{equation*}
    \Delta^{i_n} \cdots \Delta^{i_1} = \Delta^{\pi(i_n)} \cdots
    \Delta^{\pi(i_1)}
  \end{equation*}
  for every permutation $\pi$ of
  $\set{i_1,\dots,i_n}$. Thus the factorial effect defined
  in~\eqref{eq:def-interactions-weights} is symmetric in
  $i_1,\dots,i_n$ and hence the notation
  $I(\set{i_1,\dots,i_n})$ is unambiguous.
\end{remark}

\begin{remark}
  In most textbooks about factorial experiments 
  factorial effects are defined without weights, which is
  tantamount to taking
  \begin{equation*}\label{eq:weights-simple-average}
    p_{\set{i_1,\dots,i_n}}(A) = \left\{
      \begin{array}{rl}
        1 / 2^{d - n}
        &\text{ if } A \subset D \setminus \set{i_1,\dots,i_n}\\
        0 &\text{ otherwise}
      \end{array}\right.
  \end{equation*}
  in formula~\eqref{eq:def-interactions-weights}.
\end{remark}

In practice one mostly considers factorial effects of order 1 and 2.
The \emph{main 
effect} of $\set j$ is given by
\begin{equation}\label{eq:main-effect}
  I(\set{j}) = \sum_{A \subset D \setminus \{j\}}
  p_{\set{j}}(A) (\tau(A \cup \set j) - \tau(A)).
\end{equation}
It is the average effect of activating $\set{j}$'s uncertainty; it is
the average change in the output's uncertainty caused by the
activation of $\set{j}$'s uncertainty.
The \emph{interaction effect} of $\set{i,j}$ is given by
\begin{equation}\label{eq:interaction}
  I(\{i,j\}) = 
  \sum_{A \subset D \setminus \{i,j\}}
  p_{\set{i,j}}(A)(\Delta^{i}\Delta^{j}\tau)(A),
\end{equation}
where
\begin{multline*}
  (\Delta^{i}\Delta^{j}\tau)(A) =\\
  [\tau(\set{i,j} \cup A) - \tau(\set{j} \cup A)] -
  [\tau(\set{i} \cup A) - \tau(A)].
\end{multline*}
It is the average difference between the effects of activating
$\{i\}$'s uncertainty while the uncertainty of $\set{j}$'s was active
and the effect of activating $\set{i}$'s uncertainty while the
uncertainty of $\{j\}$ was inactive.

Instead of studying the activation of $\set{i}$'s uncertainty, we can
as well choose to study the inhibition of $\set{i}$'s uncertainty.  
Let
$\Delta^{\ast i} = - \Delta^{i}$ denote the ``reverse'' differential
operator, that is,
\begin{equation*}
  \Delta^{\ast i}(B) = \tau(B) - \tau(B\cup\set{i}),
\end{equation*}
and let $I^{\ast}(\set{i_1,\dots,i_n})$ denote the factorial effect of
$\set{i_1,\dots,i_n}$ obtained from $\Delta^{\ast}$, that is,
\begin{multline*}
  I^{\ast}(\set{i_1,\dots,i_n})
  =\\ \sum_{A \subset D\sm\set{i_1,\dots,i_n}}
  p_{\set{i_1,\dots,i_n}}(A)
  (\Delta^{\ast i_n} \cdots \Delta^{\ast i_1} \tau)(A).
\end{multline*}

\subsection{Example}
\label{sec:calculationslinearmodel}

Symbolic calculations of the factorial effects in a bivariate linear
model with Gaussian dependent inputs is carried out as an illustrative
example.

Let $X=(X_1,X_2)$ be normally distributed with mean 0, variance 1, and
correlation $\rho$. Let $f$ denote the function
$f(x_1,x_2)=\beta_1 x_1 + \beta_2 x_2 + \alpha x_1x_2$. Let
$\tau(A)=\expec\var(f(X)|\XcplA)$. Elementary calculations show that
$\tau(\set{2})=(1-\rho^2)(\beta_2^2+\alpha^2)$,
$\tau(\set{1})=(1-\rho^2)(\beta_1^2+\alpha^2)$ and $\tau(\set{1,2}) =              
\beta_1^2 + \beta_2^2 + \alpha^2(1+\rho^2) + 2\rho\beta_1\beta_2$.

The factorial effects are calculated as follows. Choose nonnegative weights
$p_{\set{1}}(\emptyset)$, $p_{\set{1}}(\set{2})$,
$p_{\set{2}}(\emptyset)$ and $p_{\set{2}}(\set{1})$ such that 
$p_{\set{1}}(\emptyset)+p_{\set{1}}(\set{2})=1$ and
$p_{\set{2}}(\emptyset)+p_{\set{2}}(\set{1})=1$. (We have no choice
for $p_{\set{1,2}}(\emptyset)$ which must be 1.) The effect of
activating $\set{1}$'s uncertainty while $\set{2}$'s is inactive is
given by
\begin{equation*}
  \tau(\set{1})-\tau(\emptyset) = (1-\rho^2)(\beta_1^2+\alpha^2);
\end{equation*}
the effect of activating
$\set{1}$'s uncertainty while $\set{2}$'s is active is given by
\begin{equation}
  \tau(\set{1,2})-\tau(\set{2})
  = \beta_1^2+\rho^2\beta_2^2+2\rho^2\alpha^2+
  2\rho\beta_1\beta_2.
\end{equation}
The main effect of activating
$\set{1}$'s uncertainty is then given by
\begin{multline*}
  I(\set{1}) = p_{\set{1}}(\emptyset)
  (1-\rho^2)(\beta_1^2+\alpha^2)\\ +
  p_{\set{1}}(\set{2})
  (\beta_1^2+\rho^2\beta_2^2+2\rho^2\alpha^2+
  2\rho\beta_1\beta_2).
\end{multline*}
Interchange the subscripts~1 and 2 for
$\tau(\set{2})-\tau(\emptyset)$, $\tau(\set{1,2})-\tau(\set{1})$ and
$I(\set{2})$. With equal weights, we have
\begin{multline*}
  I(\set{1}) =
  \frac{1}{2}\rho^2\beta_2^2+\frac{1}{2}(2-\rho^2)\beta_1^2 \\
  + \frac{1}{2}(1+\rho^2)\alpha^2+\rho\beta_1\beta_2.
\end{multline*}
For $I(\set{2})$ the same formula holds with the subscripts~1 and 2
interchanged.

The interaction effect between $\set{1}$ and $\set{2}$---that is, the
difference between the effect of activating the uncertainty of any
input while the uncertainty of the other is active, and, the effect of
activating that uncertainty while the uncertainty of the other is
inactive, is given by
\begin{align*}
  I(\set{1,2}) &=
                 p_{1,2}(\emptyset)(\nabla^1\nabla^2\tau)(\emptyset)\\
               &= \tau(\set{1,2})-\tau(\set{1}) -
                 [\tau(\set{2})-\tau(\emptyset)]\\
               &= \rho^2(\beta_1^2+\beta_2^2) + (3\rho^2-1)\alpha^2 +
                 2\rho\beta_1\beta_2.
\end{align*}
Here $d=2$ and hence
$I^\ast(\set{1,2})=(-1)^2(\nabla^1\nabla^2\tau)(\emptyset)=I(\set{1,2})$. 

For instance if $\rho=0$ then it holds that
$I(\set{1})=\beta_1^2+\alpha^2/2$ and $I(\set{1,2})=-\alpha^2$. The
latter means that activating any uncertainty has less effect when the
other is already active.

\section{Return to Sobol indices}
\label{sec:Sobolnewpdgm}

How can we redefine---and, therefore, understand---the family of Sobol
indices introduced in Section~\ref{sec:Intro} under the present
theory?

Let a sensitivity measure $\tau$ be chosen and \emph{define} 
\begin{eqnarray}
  \label{eq:Sobcoincidewith}
  \Sob(A) &:=& - I^{\ast}(A),\\
  \label{eq:Clocoincidewith}
  \Clo(A) &:=& \tau(D) - \tau(\cpl{A}),\text{ and},\\
  \label{eq:Totcoincidewith}
  \Tot(A) &:=& \tau(A).
\end{eqnarray}

We shall see in Proposition~\ref{prop:recoverSobolindices} below that,
under the right circumstances the set functions defined above coincide
with the usual Sobol indices. Notice that above we did \emph{not} assume that
the inputs are statistically independent.

\begin{proposition}
  \label{prop:recoverSobolindices}
  If we choose the sensitivity measure
  \begin{equation*}
    \tau(A) = \expec\var(f(X)|\XcplA)
  \end{equation*}
  and the weights
  \begin{align*}
    &p_B(A) = \left\{
      \begin{array}{rl}
        1 &\text{ if } A = D \setminus B\\
        0 &\text{ otherwise},
      \end{array}\right.
  \end{align*}
  then, under the assumption that the components of $X$ are mutually
  independent, it holds that $\Sob(A)$, $\Clo(A)$ and $\Tot(A)$
  defined in~\eqref{eq:Sobcoincidewith}, \eqref{eq:Clocoincidewith}
  and~\eqref{eq:Totcoincidewith} coincide with those same indices defined
  in Section~\ref{sec:Intro}.
  \begin{proof}
    It is trivial that~\eqref{eq:Clocoincidewith}
    and~\eqref{eq:Totcoincidewith} coincide
    with~(\ref{eq:cloindexformula}) and~(\ref{eq:totindexformula})
    under the conditions of Proposition~\ref{prop:recoverSobolindices}.
    Let us prove that~\eqref{eq:Sobcoincidewith} coincides with the index
    $\Sob(A)$ defined in the Introduction. Put $A=\set{i_1,\dots,i_n}$ and
    observe that
    \begin{multline}\label{eq:formula-diff-operator}
      \Delta^{i_n} \cdots \Delta^{i_1} \tau(B) =\\
      \sum_{B \subset C \subset B \cup A} (-1)^{|A| - |C \setminus B|} \tau(C).
    \end{multline}
    (The proof is by mathematical induction, see,
    e.g.~\cite{kojadinovic_modeling_2003} for a similar result.) 
    We now have
    \begin{align*}
      -I^{\ast}(A)
      &= -(\Delta^{\ast i_n}\cdots\Delta^{\ast i_1}\tau)(\cpl{A})\\
      &= (-1)^{n+1}(\Delta^{ i_n}\cdots\Delta^{ i_1}\tau)(\cpl{A})\\
      &= \sum_{(\cpl{A})\subset C}(-1)^{|A\sm C|}\tau(C)\\
      &= (-1)^{n+1} \sum_{B\subset A} (-1)^{|B|} \tau(D\sm B)\\
      &= (-1)^n \sum_{B\subset A} (-1)^{|B|} (\tau(D)-\tau(D\sm B))\\
      &= \sum_{B\subset A} (-1)^{|A\sm B|} \Clo(B)\\
      &= \Sob(A).
    \end{align*}
    The last equality holds because the vector of Sobol indices is the
    Möbius transform of the vector of closed indices (see,
    e.g.~\cite{mazo-tournier}). 
  \end{proof}  
\end{proposition}

Thus, Sobol indices now appear as a particular instance of the more general
theory developed in Section~\ref{sec:SensitivityMeasures}
and~\ref{sec:causes}. They appear as a particular choice of the
sensitivity measure $\tau$, the weights $p_B(A)$, and the (sign of) the
differential operator, $\Delta^i$ or $\Delta^{\ast i}$.

\section{Conclusion}
\label{sec:conclusion}

We have achieved the goal we set out to ourselves: we have carefully
redefined the Sobol indices without invoking the Sobol
decomposition. 

The journey through which we reached our
goal is far more important than the goal itself. A new viewpoint of
global sensitivity analysis has been sketched. In this new paradigm,
we need not assume statistically independent inputs; we need not limit
ourselves to the analysis of the variance. We can construct, justify,
and interpret a wide range of sensitivity indices and their
interaction effects independently of the type of index chosen. All we
need is a sensitivity measure; all we need is to construct a suitable
Dirac test. To paraphrase~\cite{ilidrissi:hal-03927476}, we have moved
away from a ``Sobol-centric'' view of sensitivity analysis.

The possibilities offered by this approach should not
obfuscate its value as a new conceptual tool allowing us to
think differently.

As a practical example, the Sobol indices now appear to follow from a
strong assumption about the weights: all the mass is put on a single
subset (remember Section~\ref{sec:Sobolnewpdgm}). Consequently, for
input $\set{j}$, the index $\Sob(\set{j})$ potentially ignores many
effects of the activation or inhibition of $\set{j}$'s uncertainty.
Putting equal mass to all subsets may appear---in the absence of any
more information or justification---as a reasonable alternative.

As another example, since the variance is no longer the only proxy for
representing the output's uncertainty, and since the variance
coincides with the choices $\rho(y,y')=(y-y')^{\alpha}/2$ and
$\alpha=2$ (remember
Section~\ref{sec:sensimeasuresexamples}), it appears that taking
$\alpha=1$ provides a reasonable alternative and may yield a 
``robust sensitivity analysis'' akin to ``robust regression''  in
statistics. 

We can unify seemingly disparate methods within the paradigm
proposed. For instance, the Shapley effect introduced in sensitivity
analysis~\cite{owen_sobol_2014} can also be recovered with a
particular choice of weights (see Appendix~\ref{sec:Shapley-effect}).

Although not covered in the present work, it is possible to estimate
sensitivity measures and their interaction effects from a Monte Carlo
sample of model runs. The methodology of~\cite{mazo-tournier} is
clearly generalizable for a wide class of Dirac tests. Moreover, since
a sensitivity measure can be identified with a symmetric factorial
experiment with two levels, we can look at that literature.

Optimizing the tradeoff between the estimator's performance and the
number of model runs is another challenge.  For this, once again, we
could look at the appropriate literature, especially that about
fractional factorial experiments. We may also want to exploit certain
properties of certain sensitivity measures (e.g., monotonicity). If
the inputs are statistically dependent then a yet another challenge is
to be able to sample from the conditional distributions.

Beyond sensitivity analysis, artificial intelligence and machine
learning models  explain their predictions with indices similar
to sensitivity indices~\cite{ verdinelli_feature_2024,
  bilodeau_impossibility_2024, strumbelj_efficient_2010}.  Although
the specificity of those models and the contexts in which they are
used will have to be taken into account (the output is no longer a
function of the inputs; the definition of $\tau$ must change), the
paradigm proposed may serve as a basis for thinking those indices
as well.

Finally and intriguingly, sensitivity indices so far have been mainly
studied as association measures between an output $Y$ and an input
$X_j$. It is then desirable that the association measure is zero if
and only if $X_j$ and $Y$ are independent, and one if and only if $Y$
is a function of $X_j$. Most sensitivity indices based on probability
measures also follow this philosophy: they compare the distribution of
$Y$ with that of $Y$ conditionally on $X_j$. See,
e.g.~\cite{borgonovo_convexity_2025}. We took a different approach: we
compare the output's conditional distribution not with the
unconditional one, but with the Dirac at $X_{\cpl{\set{j}}}$. It would be
interesting to better understand the connection between these two
``dual'' approaches.

\appendix

\section{The Sobol decomposition and the indices that ensue}
\label{sec:SobolIndices}

The Sobol  decomposition is the following
result~\cite{sobol1993sensitivity,vaart1998asymptotic}. 

\begin{proposition}[Sobol decomposition]
  \label{prop:SHdecomposition}
  If $X$ is random vector in $\real^d$ with independent components and
  $f$ is a real function defined on $\real^d$ such that
  $\expec f(X)^2 < \infty$ then there are (almost surely) unique
  functions $s_A$, $A\subset D$, such that
  $f(X) = \sum_{A\subset D} s_A(X_A)$ with $s_\emptyset = \expec f(X)$
  and $\expec( s_A( X_A) | X_B ) = 0$ if $A \not \subset B$.
\end{proposition}

A corollary is that the covariance of every pair of the random
variables $s_A(X_A)$, $A\subset D$, is null, and hence
\begin{equation*}
  \var f(X) = \sum_{A\subset D} \var s_A(X_A).
\end{equation*}
The number
$\var s_A(X_A) =: \Sob(A)$
is called the Sobol index  associated with $A\subset D$ and by
definition we have
\begin{equation*}
  \Clo(A) = \sum_{B\subset A}\Sob(B),\quad
  \Tot(A) = \sum_{B\cap A\ne\emptyset}\Sob(B).
\end{equation*}

\section{Shapley effects}
\label{sec:Shapley-effect}

Let $\Clo(A)$ denote the closed Sobol index as defined in the
Introduction. The Shapley effect of input $\set{j}$ ($j=1,\dots,d$) is
defined as
\begin{equation*}
  \Shap(\{j\}) = \frac{1}{d} \sum_{A\subset D\setminus \{j\}}
  \frac{1}{\binom{d-1}{\lvert A\rvert}}
  (\Clo(A\cup\set{j}) - \Clo(A)).
\end{equation*}
See~\cite{owen_sobol_2014}. Borrowed from cooperative game
theory~\cite{owen_game_1982}, the Shapley effect of
input $\set{j}$ represents its ``fair contribution'' should certain
axioms be imposed. Substituting $\Tot(A)$ for $\Clo(A)$ does not
change the value of the index. See~\cite{da_veiga_basics_2021} for
details. 

\begin{proposition}
  \label{prop:recoverShapley}
  If we choose the sensitivity measure $\tau(A)=\expec\var(f(X)|\XcplA)$
  and the weights
\begin{align}\label{eq:weights-Shapley}
  &p_B(A) = \left\{
    \begin{array}{rl}
      \frac{1}{(|D \setminus B|+1) \binom{|D \setminus B|}{|A|}}
      &\text{ if } A \subset D \setminus B\\
      0 &\text{ otherwise},
    \end{array}\right.
\end{align}
then $I(\set{i_1})$ defined in~(\ref{eq:main-effect}) coincides with
$\Shap(\set{i_1})$.
\begin{proof}
  The proof follows from~(\ref{eq:formula-diff-operator}).
\end{proof}
\end{proposition}

If $\tau$ were to represent a game~\cite{owen_game_1982} then
$I(\set{i_1,\dots,i_n})$ with the weights~(\ref{eq:weights-Shapley})
would coincide with the Shapley interaction
index~\cite{Grabisch1997Alternative,grabisch_axiomatic_1999}.

\section*{Acknowledgments}
The author thanks an anonymous reviewer for their detailed and
constructive comments which helped to improve the manuscript.  The
author thanks Laurent Tournier for his critical reading of the
manuscript which helped to improve its presentation.

\section*{Supplement}
``pdf'' file with the proof of Theorem~\ref{thm:fundamentaltheorem}.

\bibliographystyle{abbrv}
\bibliography{biblio.bib}

\end{document}

\maketitle

%%%%%%%%%%%%%%%%%%%%%%%%%%%%%%%%%%%%%%%%%%%%%
%% MAIN TEXT %%
%%%%%%%%%%%%%%%%%%%%%%%%%%%%%%%%%%%%%%%%%%%%%

Let $f$ be a Borel function from the $d$-dimensional Euclidean space
$\real^d$ into the real line $\real$.
Let $P$ be a probability measure on $\real^d$ 
equipped with its Borel $\sigma$-field $\Borel^d$. Let $X$ be a
random vector in $\real^d$ with distribution $P$, denoted by $X\sim P$.
Denote by $\Omega$  the probability space on which $X$ is
defined. Probabilities on $\Omega$ will be denoted with the symbol
$\Pr$.  
Denote $D=\{1,\dots,d\}$. If $A\subset D$ then
$X_A\in\real^{|A|}$ denotes the subvector of $X$ with components those
of $X$ indexed by $A$. For instance if $d=4$ and $A=\{3,1,4\}$ then
$X_A = (X_1,X_3,X_4)$. If $A=D$ then $X_D=X$. If $A=\emptyset$ then
$X_{\emptyset}$ denotes an arbitrary constant. Let $\powerset$ denote
the power set of $D$, that is, the set of all of its subsets.

We say that a set $A\subset D$ is \emph{superfluous} if there is a
Borel function $g$ such that $f(X) = g(X_{\cpl A})$ almost surely.  In
this case we also say that $f$ is \emph{unsensitive} to
$A$. Otherwise, $A$ is said to be \emph{non-superfluous}, and $f$
\emph{sensitive} to $A$.  Note that $\emptyset$ is superfluous, and
that $D$ is non-superfluous unless $f$ is constant. Note also that
$g(\XcplA)$ must be a version of the conditional expectation of $f(X)$
given $\XcplA$, denoted by $\expec(f(X)|\XcplA)$. Finally note that if
$A\subset B$ and $B$ is superfluous then so must be $A$.

\begin{definition}[sensitivity measure]\label{def:sensimeasu}
  A mapping $\tau$ of $\powerset$ into $\real$ is said to be a
  sensitivity measure if it satisfies the two following axioms:
  \begin{enumerate}[(i)]
  \item \label{enum:sensimeasu:nonnegative}
    $\tau$ is nonnegative;
  \item \label{enum:sensimeasu:zero}
    $\tau(A) = 0$ if and only if $A$ is superfluous.
  \end{enumerate}
\end{definition}

The number $\tau(A)$ measures the sensitivity of $f$ to its inputs
indexed by $A$. The number $\tau(D)$ is called \emph{the total
  sensitivity of $f$}, or simply \emph{the sensitivity of $f$} and
measures the uncertainty about the output of the function $f$.

\begin{lemma}
  \label{lem:technicallem}
  The following statements are equivalent:
  \begin{enumerate}[(i)]
  \item \label{enum:technicallem:superfl}
    $A$ is superfluous;
  \item \label{enum:technicallem:conditioning}
    $\Pr\{f(X)=g(\XcplA)|X_B|\}=1$ almost surely for every
    $B\subset D$;
  \end{enumerate}
  
  \begin{proof}[Proof of Lemma~\ref{lem:technicallem}]
    The proof that~\eqref{enum:technicallem:superfl}
    implies~\eqref{enum:technicallem:conditioning} follows from the equality
    \begin{equation*}
      0 = \expec \Pr\{f(X)\ne g(\XcplA)|X_B\}.
    \end{equation*}
    For the converse take expectations.
  \end{proof}
\end{lemma}

We say that a functional $\phi(Q)$ of a probability measure $Q$ is a
\emph{Dirac test} if $\phi(Q)\ge 0$ and $\phi(Q) = 0$ if and only if
$Q$ is a Dirac measure. Here $Q$ is assumed to be a probability
measure on $\real$ equipped with its Borel $\sigma$-field.

Fix a subset $A$ of $D$ and denote by $P_t$ the probability measure
$\Pr\{X\in\cdot|\XcplA=t\}$. Let $P_t\circ f^{-1}$ denote the
probability measure $\Pr\{f(X)\in\cdot|\XcplA=t\}$.

\begin{lemma}
  \label{lem:Diractestisnul}
  It holds that $\phi(P_t\circ f^{-1}) = 0$ if and only if
  the function $f$ is conditionally constant given $\XcplA=t$, that
  is, for some real $y$, we have $\Pr\{f(X)=y|\XcplA=t\}=1$.

  \begin{proof}[Proof of Lemma~\ref{lem:Diractestisnul}]
    If $\phi({P}_t\circ f^{-1}) = 0$ then by definition
    there is $y$ such that
    ${P}_t\circ f^{-1}(H) = \delta_y(H)$ for every Borel set
    $H$, where $\delta$ denotes the Dirac measure. In particular with
    $H=\set{y}$ we get
    ${P}_t\circ f^{-1}(\real\sm\set{y}) = 0$.
    But this means exactly that
    $\Pr\{f(X)=y|\XcplA=t\}=1$. Conversely, if the latter holds for
    some $y$ then so does
    ${P}_t\circ f^{-1}(\real\sm\set{y}) = 0$. Choose $H$ a
    Borel set in $\Borel$. If $H$ contains $y$ then $\real\sm H$ is
    contained in $\real\sm\set{y}$ and hence
    ${P}_t\circ f^{-1}(H) = 1$. A similar reasoning yields
    that ${P}_t\circ f^{-1}(H) = 0$ if $H$ does not contain $y$.
    Therefore, $P_t\circ f^{-1}$ is indeed a Dirac measure.
  \end{proof}
\end{lemma}

Now we allow $A$ to vary and define
\begin{equation*}
  \phi_A(t) = \phi(\Pr\{f({X})\in\cdot|\XcplA=t\})
  \quad (t\in\real^{|\cpl{A}|}).
\end{equation*}

\begin{theorem}
  \label{thm:quasisensimeasu}
  The mapping defined by
  \begin{align}
    \tau(A) &= \expec \phi_A(\XcplA)
              \notag\\
            &= \int_{\real^{|\cpl{A}|}}
              \phi_A(t)\Pr\{\XcplA\in\diff t\}
              \notag\\
            &= \int_{\real^{|\cpl{A}|}}
              \phi(\Pr\{f({X})\in\cdot|\XcplA=t\})
              \Pr\{\XcplA\in\diff t\}
              \label{eq:quasisensimeasu}
  \end{align}
  is a sensitivity measure.
  \begin{proof}[Proof of Theorem~\ref{thm:quasisensimeasu}]
    Suppose that $\tau(A)=0$. From Lemma~\ref{lem:Diractestisnul}, it
    holds that for almost all $\omega$ there is some $Y^\omega$ such
    that $\Pr\{f(X)=Y^\omega|\XcplA\}_\omega$, the value of the
    conditional probability at $\omega$, is equal to 1. (Note that the
    event in the probability is the event $\{\omega'\in\Omega:
    f(X_A(\omega'),\XcplA^\omega)=Y^\omega\}$.) To show that $A$ is
    superfluous, it remains to show that the induced mapping
    $Y^\omega$ of $\omega$ is measurable $\XcplA$. That is
    true. Indeed, denote
    $E_\omega=\set{s\in\real^{|A|}:(s,\XcplA^\omega)\in
      f^{-1}(Y^\omega)}$ and observe that
    \begin{align*}
      Y^\omega &= Y^\omega
                 \Pr\{X_A\in E_\omega|\XcplA\}_\omega\\
               &= \int_{E_\omega} Y^\omega
                 \Pr\{X_A\in\diff s|\XcplA\}_\omega\\
               &= \int_{E_\omega} f(s,\XcplA^\omega)
                 \Pr\{X_A\in\diff s|\XcplA\}_\omega
    \end{align*}
    is a function of $\XcplA^\omega$.

    Conversely, if $A$ is superfluous, then there is a function
    $g(\XcplA)$ such that $\Pr\{f(X)=g(\XcplA)\}$ and hence
    $\Pr\{f(X)=g(\XcplA)|\XcplA\}$ is equal to 1 almost surely. From
    Lemma~\ref{lem:Diractestisnul} this implies that
    \begin{equation*}
      \phi(\Pr\{f({X})\in\cdot|\XcplA=t\}) = 0
    \end{equation*}
    almost everywhere $P$, and hence $\tau(A)=0$.
  \end{proof}
\end{theorem}